\documentclass[10pt,twocolumn,letterpaper]{article}

\usepackage{wacv}
\usepackage{times}
\usepackage{epsfig}
\usepackage{epstopdf} %converting to PDF
\usepackage{graphicx}
\usepackage{amsmath}
\usepackage{amssymb}
\usepackage{subcaption}

\wacvfinalcopy % *** Uncomment this line for the final submission

\def\wacvPaperID{****} % *** Enter the wacv Paper ID here

\ifwacvfinal\fi
\begin{document}

%%%%%%%%% TITLE
\title{Thermal to Visible Synthesis of Face Images using Multiple Regions}

% Authors at the same institution
\author{Benjamin S. Riggan\textsuperscript{1,*} \hspace{2cm} Nathaniel J. Short\textsuperscript{1,2} \hspace{2cm} Shuowen Hu\textsuperscript{1} \\\\
\textsuperscript{1}U.S. Army Research Laboratory, 2800 Powder Mill Rd., Adelphi, MD 20783\\
\textsuperscript{2}Booz Allen Hamilton, 8283 Grennsboro Dr., McLean, VA 22102\\
{\tt\small \textsuperscript{*}Corresponding author: benjamin.s.riggan.civ@mail.mil}
}
% Authors at different institutions
%\author{Benjamin S. Riggan \\
%U.S. Army Research Laboratory\\
%{\tt\small benjamin.s.riggan.civ@mail.mil}
%\and
%Nathaniel J. Short \\
%Booz Allen Hamilton\\
%{\tt\small Short_Nathaniel@bah.com}
%\and
%Shuowen Hu \\
%U.S. Army Research Laboratory\\
%{\tt\small shuowen.hu.civ@mail.mil}
%}

\maketitle
\ifwacvfinal\thispagestyle{empty}\fi

%%%%%%%%% ABSTRACT
\begin{abstract}
%   Synthesis of visible spectrum faces from polarimetric thermal faces is beneficial for providing automatic face recognition capabilities (e.g., landmark detection and matching) at night and for enabling analysts to verify the identification results.
Synthesis of visible spectrum faces from thermal facial imagery is a promising approach for heterogeneous face recognition; enabling existing face recognition software trained on visible imagery to be leveraged, and allowing human analysts to verify cross-spectrum matches more effectively.
We propose a new synthesis method to enhance the discriminative quality of synthesized visible face imagery by leveraging both global (e.g., entire face) and local regions (e.g., eyes, nose, and mouth).  Here, each region provides (1) an independent representation for the corresponding area, and (2) additional regularization terms, which impact the overall quality of synthesized images.   We analyze the effects of using multiple regions to synthesize a visible face image from a thermal face.  We demonstrate that our approach improves cross-spectrum verification rates over recently published synthesis approaches.  Moreover, using our synthesized imagery, we report the results on facial landmark detection---commonly used for image registration---which is a critical part of the face recognition process.   
%   The proposed method provides improved synthesized visible imagery compared to recent thermal-to-visible synthesis work. We show that proposed synthesis approach achieves superior visual similarity to corresponding visible images compared to previous methods. More importantly, we demonstrate that our synthesized images are interoperable with existing commercial, government, and open source face detection and recognition software.
\end{abstract}

%%%%%%%%% BODY TEXT
\section{Introduction}

Commercial-off-the-shelf (COTS), government-off-the- shelf (GOTS), and open-source face recognition software are widely used for automatically searching large galleries of subjects, with a face image of an unknown subject, to identify a smaller subset of potential candidates.
%downselecting large galleries of subjects to a manageable number of likely candidates. 
These candidate matches are commonly presented to expert analysts for human adjudication.   Almost all deployed facial recognition software, whether for commercial, law enforcement, or defense related applications, operate in the visible bands of the electromagnetic spectrum due to the ubiquity of low cost visible cameras.
%Most face recognition systems only consider visible spectrum imagery since visible cameras are ubiquitous in surveillance/security systems.

The ability to synthesize a visible spectrum face from facial imagery acquired in the infrared spectrum provides a unique way to expand conventional face recognition capabilities beyond the visible spectrum without the need to develop highly customized software. Therefore, we propose a new synthesis method that leverages both global and local regions to produce a visible-like face image from a thermal infrared face image, so that existing face detection, landmark detection, and face recognition algorithms may be directly applied.

%Commercial-off-the-shelf (COTS), government-off-the- shelf (GOTS), and open-source face recognition systems are widely used for automatically down selecting large galleries of subjects to a manageable number of likely candidates. These candidate matches are commonly presented to expert analysts for human adjudication. Most face recognition systems only consider visible spectrum imagery since visible cameras are ubiquitous in surveillance/security systems.

\begin{figure}[tbp]
  \centering
  \begin{subfigure}[b]{0.65\linewidth}
    \includegraphics[width=\linewidth]{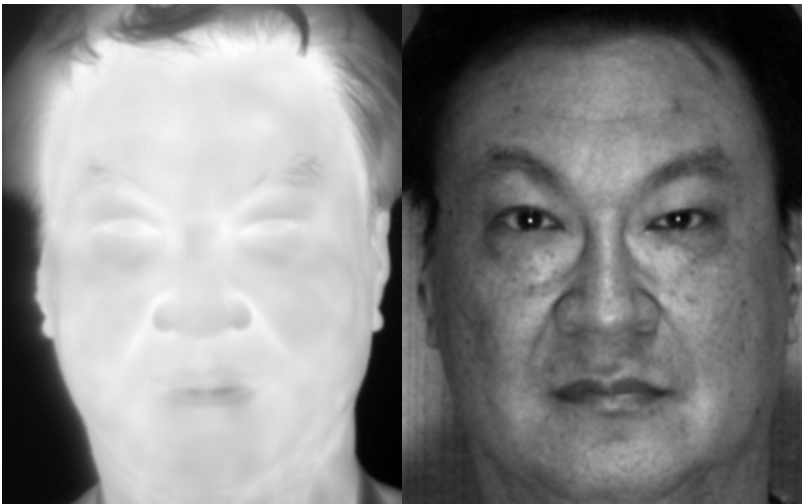}\\
    \caption{}	
  \end{subfigure}\\
 \begin{subfigure}[b]{0.65\linewidth}
    \includegraphics[width=\linewidth]{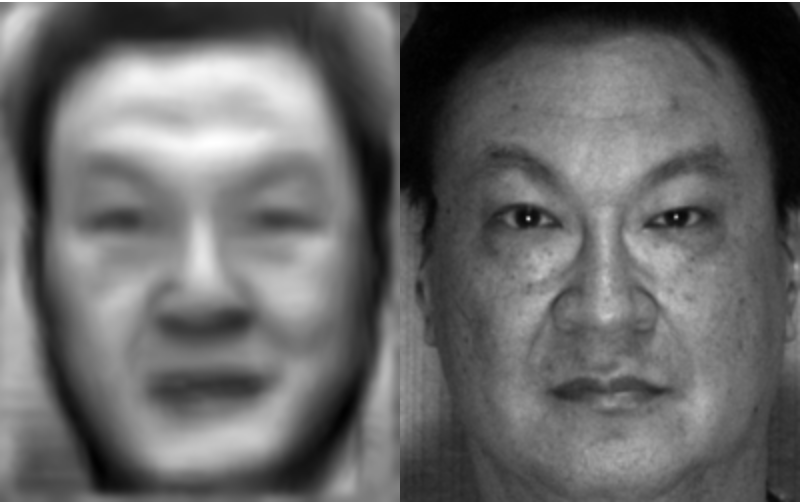}\\
    \caption{}	
  \end{subfigure}
  \caption{(a) Thermal infrared and visible images of a subject demonstrate the large modality gap, which make matching and adjudication much more challenging.  Whereas, (b) synthesized visible (from a thermal infrared) and visible images allow for more effective cross-spectrum matching and adjudication.}
  \label{fig:challenge}
\end{figure}

%There has been a lot work performed recently on visible spectrum face recognition. 
Due to the abundance of visible spectrum face imagery (via social media, CCTV, etc.), improvements in processing hardware (CPU, GPU, and memory speeds), and advancement of machine learning algorithms, deep neural networks \cite{ParkhiVedaldi2015} have attained significant improvement in face recognition performance. Deep neural networks, such as convolutional neural networks (CNNs), have demonstrated near human-level recognition in the visible spectrum \cite{TaigmanYang2014}. CNNs are able to learn deep hierarchical feature representations that are robust to natural variations, including but not limited to pose, illumination, and expression (PIE) conditions.

Attention to infrared spectrum face recognition has increased over the last several years \cite{BourlaiKalka2010, JuefeiXuPal2015, KlareJain2010, NicoloSchmid2012}. Thermal infrared imaging is ideal for covert nighttime face recognition. However, thermal face signatures acquired at night need to be compared with visible face signatures from existing face databases and watchlists. Thermal-to-visible face recognition is challenged by a substantial modality gap between signatures in each domain. Recent approaches on cross-spectrum face recognition have partially addressed this issue, and it has been shown that incorporating polarization state information reduces the modality gap even further \cite{ShortHu2015}.  In this paper, thermal imaging may refer to either conventional thermal or polarimetric thermal imaging.

Although state-of-the-art cross-spectrum face recognition algorithms have demonstrated significant promise \cite{GurtonYuffa2014,ShortHu2015b,ShortHu2015}, these customized algorithms are difficult to integrate into existing infrastructure and common practices. The top matches from recent cross-spectrum face recognition methods are not easily verifiable by human analysts due to the large differences in visual appearance of facial imagery collected in the thermal and visible bands of the electromagnetic spectrum. Moreover, a significant amount of time and resources have been invested in developing existing visible face recognition technology, which are less effective on cross-spectrum face recognition tasks.  Thus, there is a need to alleviate the difficulty of matching thermal and visible spectrum imagery.  

An example that demonstrates how thermal-to-visible synthesis can be used facilitate cross-spectrum matching is shown in Fig.~\ref{fig:challenge}.

Therefore, the primary objective of this is work is to present a new method for synthesizing visible spectrum face imagery from conventional thermal or polarimetric thermal face images.  This method exploits the complementary representations and effects from various regularizations that are induced by multiple regions on the resulting synthesized image.   We compare recognition performance of our approach with existing synthesis methods, and we also explore using the synthesized imagery for landmark detection.

%(1) implementing a multi-region cross-spectrum synthesis method, (2) demonstrating improved quality of synthesized imagery, and (3) demonstrating effectiveness of using synthesized imagery for detecting fiducial landmarks and for face recognition (using COTS and pre-trained CNNs).
%
%We show that our proposed method achieves higher recognition rates than existing cross-spectrum synthesis approaches.

\begin{figure*}[htbp]
  \centering
  \includegraphics[width=0.9\textwidth]{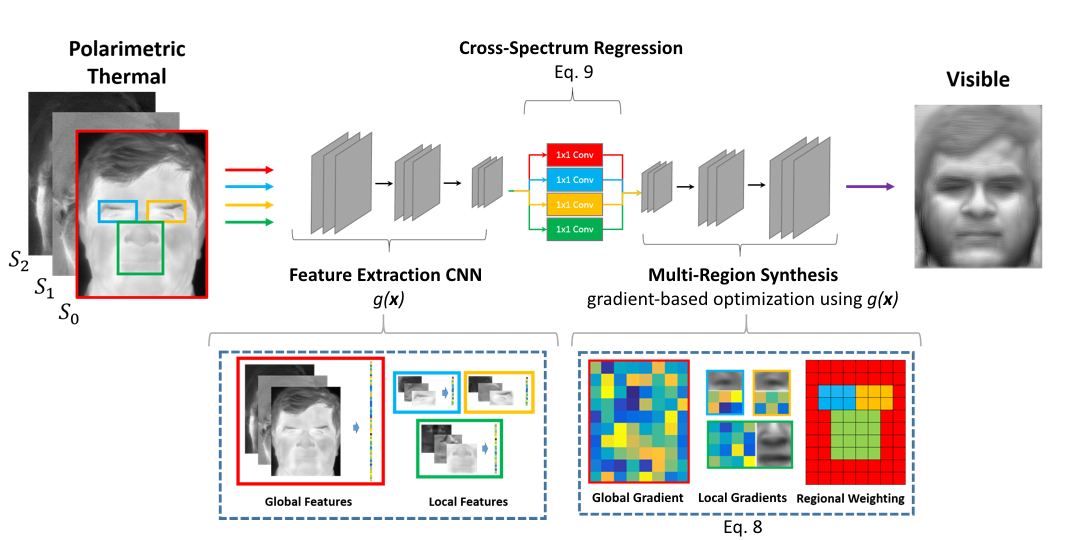}
  \caption{Given a thermal image, features are first extracted from a global region of interest (red) and local fiducial regions of interest (blue, yellow, and green) using a fully convolutional neural net, $g(\cdot)$.  Then, region specific cross-spectrum mappings are used to estimate corresponding visible representations from the extracted thermal features.  Lastly, by backpropagating the error between $g(\mathbf{x})$ and the estimated visible features from each region, we determine the gradient updates from both global and local regions, which are combined in order to produce a synthesized visible image.}
\label{fig:overview}
\end{figure*}

%-------------------------------------------------------------------------
\section{Cross-Spectrum Face Recognition \& Synthesis}

The recent interest in cross-spectrum face recognition has led to two types of approaches: (1) custom cross-spectrum face recognition, and (2) cross-spectrum synthesis.  The first approach is similar to traditional visible-to-visible face recognition.  In general, traditional face recognition algorithms leverage a mapping that transforms images (commonly from a single spectral band) to a corresponding latent feature representation, which can then be matched with latent representations of gallery images either directly using similarity/dissimilarity metrics (e.g., cosine similarity, Euclidean distance) or indirectly using classifiers, such as partial least squares, support vector machines, or softmax classifiers.  However, cross-spectrum face recognition aims to map corresponding images from different domains (e.g., visible and thermal) to a common latent subspace, so that matching can be performed.
%Fundamentally, facial biometrics problems may be largely separated into two (very much related) types: recognition and synthesis. Furthermore, these two types of problems can be subdivided into within-spectrum or cross- spectrum. In this section, we present a brief summary of each of these four problems.
%\subsection{Recognition}
%\subsubsection{Within-Spectrum}
%
%Typically, visible-to-visible face recognition is implied when referring to within-spectrum face recognition. While there are some studies focusing on infrared-to-infrared face recognition \cite{Wolff2005}, it is significantly less likely to have enrolled infrared signatures of persons-of-interest as a reference. Most galleries (or watch lists) contain visible spectrum signatures of individuals faces.
%
%Classic challenges for within-spectrum matching include: pose, expression, illumination, and low-resolution. Recently, visible-to-visible face recognition has extended beyond controlled laboratory conditions \cite{HuangJain2007,HuangMattar2012} , where web imagery \cite{YiLei2014} is being exploited to provide enhanced recognition capabilities. Also, there has been work on applying domain adaptation and transfer learning \cite{RenDai2014} for within-spectrum face recognition in order to account for differences between datasets (e.g., sensors/cameras, perspectives, etc.).
%
%\subsubsection{Cross-Spectrum}

In \cite{HuChoi2015}, a one-vs-all framework using binary classifiers, such as partial least squares (PLS), was used for thermal-to-visible face recognition, where thermal counter examples were used to augment the negative sample set during training.  This approach demonstrated improved cross-spectrum recognition performance.  However, there is a relatively fast rate of diminishing returns as the number of counter examples increases.

%In \cite{RigganShort2016b}, multilayer neural networks were used to learn correlated feature representations for thermal-to-visible face recognition. 
Sarfraz et al. \cite{SarfrazStiefelhagen2015} trained a model to estimate local thermal features (e.g., HOG or SIFT) from visible features. The reverse, namely estimating visible features from thermal features, was also considered, but a slight performance drop was reported. Riggan et al. \cite{RigganReale2015} proposed an auto-associative neural network model that learns a common latent subspace in which the visible and thermal features are highly correlated, so that a classifier trained using visible imagery generalizes to thermal imagery.

More recently, \cite{RigganShort2016} combined feature regression methods and PLS for polarimetric thermal-to-visible face recognition. This approach proposed a framework that combined local regression neural nets (e.g., \cite{SarfrazStiefelhagen2015, RigganReale2015}) with a discriminative classifier, outperforming state-of-the-art methods for both conventional and polarimetric thermal-to-visible face recognition.

%\subsection{Synthesis}
The second class of methods, namely synthesis-based methods, perform image generation or image visualization.  Here, the objective is to generate an image based on some conditional information (e.g., a latent representation, an image, or a class label).  Synthesis-based methods were first studied for visible images.  For example, Hoggles \cite{VondrickKhosla2013} generated images based on Histogram of Oriented Gradient (HOG) representation in order to better understand detection failures, and \cite{MahendranVedaldi2015} demonstrated improved synthesis results using HOG, dense scale invariant feature transform (DSIFT), and CNN features.  Manhendran et al.~\cite{MahendranVedaldi2015} pose the problem as the following optimization problem:
\begin{equation}
\mathbf{x}^* = \arg\min_{\mathbf{x}} \ell(f(\mathbf{x}),\mathbf{y}) + \lambda R(\mathbf{x})
\end{equation}
where $\mathbf{y}$ denotes the given feature vector, $f(\mathbf{x})$ is function that extract features from a given image $\mathbf{x}$, $\ell(\cdot)=\|f(\mathbf{x})-\mathbf{y}\|^2$, and $R(\mathbf{x})$ denotes the regularization term.   The regularization term encourages (1) the pixels intensities to lie in a specific range using $\|\mathbf{x}\|^\alpha_\alpha$, and (2) a piece-wise constant reconstruction using 
a total variation constraint $\sum_{i,j} \left ( \left ( x_{i,j+1} - x_{i,j} \right )^2 + \left ( x_{i+1,j} - x_{i,j} \right )^2\right )^{\beta / 2}$

Specifically for cross-spectrum synthesis, the goal is to synthesize a corresponding visible image from a given thermal image (or polarimetric thermal image stack).  The main reasons for performing cross-spectrum synthesis are: (1) to enable easy verification by a human-in-the-loop and (2) 
%to exploit recent state-of-the-art advances in visible face recognition. 
to provide direct integration with state-of-the-art commercial and academic face recognition algorithms aimed at matching visible faces.

The first method to perform thermal-to-visible and polarimetric thermal-to-visible synthesis was Riggan et al.~\cite{RigganShort2016b}.   This study demonstrated the initial success for synthesizing a visible image from a thermal face image, reporting approximately a 20\% improvement in structural similarity on average.  For this approach, the following optimization performed:
\begin{equation}
\label{eq:opt2}
\mathbf{x}^* = \arg\min_{\mathbf{x}} \ell(g(\mathbf{x}), h \circ g(\mathbf{t})) + \lambda R(\mathbf{x})
\end{equation}
where, like \cite{MahendranVedaldi2015}, the objective is to determine the optimal image $\mathbf{x}$ that produces similar features using a fixed mapping $g(\cdot)$.  However, in \cite{RigganShort2016b} the given features are produced by $g(\cdot)$ followed by another function $h(\cdot)$, which maps thermal (or polarimetric thermal) features to corresponding visible features.  Since the mapping $g$ is fixed and the $h$ is pretrained, the optimal visible image representation is given by solving Eq.~\ref{eq:opt2} when given a thermal image.

Also, \cite{ZhangPatel2017} applied a conditional generative adversarial networks (CGAN) to perform cross-spectrum synthesis, which extends upon GAN-based methods \cite{GoodfellowPouget2014,RadfordMetz2015}. This state-of-the-art method was able to produce images with photo-realistic texture and report face verification results in the form of true positive rate versus false positive rate.  The discriminability of the synthesized imagery in \cite{ZhangPatel2017}, though surpassing the state-of-the-art at that time, can still be further improved.
%However, in comparison with custom cross-spectrum recognition methods, the verification rates are somewhat lacking.  Our hypothesis is that some amount of discriminability may be sacrificed when producing a photo-realistic image.
 
The synthesis problem is fundamentally related to the recognition problem. In principle, if one can generate an image to be sufficiently discriminative, then one could inherently do recognition.

\section{Cross-Spectrum Synthesis using Multiple Regions}
The primary objective is to synthesize a visible-like image from a given polarimetric thermal or conventional thermal face image.  In practice, a polarimetric thermal image is composed of three Stokes vectors (or images): $S_0$, $S_1$, and $S_2$, where $S_0$ is the total intensity image (i.e.,  conventional thermal image), $S_1$ is the difference between horizontal and vertical polarization states, and $S_2$ is the difference between diagonal (45 degree and 135 degree) polarization states. Additionally, the degree of linear polarization (DoLP) representation can be derived from these Stokes vectors: $I_{DoLP}=(S_1^2 + S_2^2)^{1/2} / S_0$ \cite{YuffaGurton2014}.  The polarization state information is useful in providing additional information about the structure and geometry of faces that is not provided by conventional thermal imagery.

Here, we present a cross-spectrum synthesis method that enhances facial details (compared to recent work \cite{RigganShort2016b,ZhangPatel2017}) by optimizing an objective function that comprises complementary representations and regularizations that are induced from multiple fiducial regions. This multi-region objective function jointly leverages both global and local evidences to generate images that preserve overall facial structure and local fiducial details.  An overview of our approach is shown in Fig.~\ref{fig:overview}.

\subsection{Problem Formulation}
Consider a given thermal face image, $\mathbf{t} \in  \mathbb{R}^{h \times w \times c}$, where the goal is to produce an image $\mathbf{x} \in \mathbb{R}^{h \times w}$ that is sufficiently similar to the corresponding visible image representation of $\mathbf{t}$.  Let the images be indexed by $0 < u < h-1$ and $0 < v < w-1$, where $\mathbf{t}_{u,v}$ and $x_{u,v}$ denotes the intensity value (or vector) at the pixel location $(u,v)$.  Note that $h$, $w$, and $c$ denote the height, width, and number of channels for the images, respectively.  In general, $\mathbf{t}$ may be either a conventional thermal image (i.e., $c=1$) or a polarimertric thermal image (i.e., $c>1$) that may use any combination of $S_0$, $S_1$, $S_2$ (including DoLP).

In this work, we consider the impact of multiple loss and regularization functions that are induced by predefined regions of interest (ROIs).   Although, these ROIs are arbitrary, we consider ones corresponding to local discriminative features (e.g., eyes, nose, and mouth) and one global region that contains the entire face.   

%\subsection{Optimization}
%The proposed method is conceptually similar to \cite{RigganShort2016b}, where the input is a thermal image, $\mathbf{t} \in \mathbb{R}^{h \times w \time s c}$, and the synthesized output is an image $\mathbf{x} \in \mathbb{R}^{h \times w}$.  Here, $h$, $w$, and $c$ denote the height, width, and number of channels for the images, respectively.  
%The input thermal may be a conventional thermal image (i.e., $c=1$) or a polarimetric thermal image (i.e., $c>1$).  However, unlike \cite{RigganShort2016b}, we are able to produce improved synthesized imagery by seamlessly optimizing over multiple regions of interest. For example, these regions may correspond to facial areas of interest, such as eyes, nose, mouth, or even a tight crop region around these facial features.
 
Consider a set of ROIs, $\left \{\beta_1, \dots, \beta_K \right \}$, where $\beta_i$ represents the $i^{th}$ ROI within a given thermal image $\mathbf{t}$.  For each region, we want to minimize the following objective function
\begin{equation}
J_i(\mathbf{x}) =\left\{\begin{matrix}
  \mathcal{L}(\tilde{\mathbf{x}}_i) &,\; \tilde{\mathbf{x}}_i = \{ x_{u,v}  : (u,v) \in \beta_i \}\\ 
 0 &,\; otherwise
\end{matrix}\right. ,
\end{equation}
where $\mathcal{L}(\tilde{\mathbf{x}}_i)=\ell(g(\tilde{\mathbf{x}}_i), h_i \circ g(\tilde{\mathbf{t}}_i)) + \lambda R(\tilde{\mathbf{x}}_i)$ represents the objective function similar to \cite{RigganShort2016b}, except it is computed using $\tilde{\mathbf{x}}_i$ and $\tilde{\mathbf{t}}_i$, which are defined over the region $\beta_i$.  Here, $\ell(\mathbf{a},\mathbf{b})=\left \| \mathbf{a} - \mathbf{b} \right \|^2$ is the loss function, $R(\mathbf{x})$ enforces the $\alpha$-norm and total variation penalties, $g(\cdot)$ represents a generic mapping from an input image to some representative features, and $h_i(\cdot)$ is a region specific cross-spectrum mapping that corresponds to the region $\beta_i$.  In practice, we implement $g$ as a fully convolutional neural network so that the size of the input image does not have to be defined, and $h_i$ is composed of $1\times1$ convolutional layers.

A synthesized image is generated by solving the following optimization problem:
\begin{equation}
\mathbf{x}^* = \arg\min_{\mathbf{x}} J(\mathbf{x}),
\label{eq:opt}
\end{equation}
where
\begin{equation}
J(\mathbf{x}) = \sum_i \omega_i J_i(\mathbf{x}). 
\label{eq:total_obj}
\end{equation}
The combined objective function in Eq.~\ref{eq:total_obj} tries to balance global structure and local details using the weights, $\omega_i$, corresponding to each region.  Note that $\sum_i \omega_i = 1$, which implies that there must be at least one of the $K$ regional objectives enforced.
%From a thermal imag $\mathbf{t}$, we can extract a set of images $\left \{ \mathbf{t}_1, \dots, \mathbf{t}_K \right \}$, where $\mathbf{t}_i \subseteq \mathbf{t}$  corresponding to the the $i^{th}$ region.  Then, we synthesize a visible-like image by minimizing the following objective function:
%\begin{equation}
%J(\Theta) = \sum_i w_i J_i(\Theta), 
%\label{eq:total_obj}
%\end{equation}
%where
%\begin{equation}
%J_i = \ell(f(\mathbf{x}_i), g_i(\mathbf{t}_i)) + \lambda R(\mathbf{x_i})
%\label{eq:obj_i}
%\end{equation}
%In Eq.~\ref{eq:total_obj}, $w_i$ denotes the weight associated with the $i^{th}$ region.  In Eq.~\ref{eq:obj_i}, $\ell(\cdot)=\left \| f(\mathbf{x}) - g(\mathbf{t}) \right \|$, $R(\mathbf{x}_i)$ represents the $\alpha$-norm and total variation regularization penalties (see \cite{ZeilerFergus2014}) on the synthesized image region $\mathbf{x}_i$, and the mappings $f(\mathbf{x})$ and $g_i(\mathbf{t})$ represent the visible and thermal feature extraction models.

The multi-region objective generalizes the approach in \cite{RigganShort2016b}.   When $K=1$, Eq.~\ref{eq:total_obj} reduces to the objective in Eq.~\ref{eq:opt2}.  However, when $K > 1$, the generalized multi-region objective function leverages multiple overlapping regional objectives that can be combined to synthesize a face image with enhanced discriminative facial features.

Similar to \cite{MahendranVedaldi2015}, we use gradient descent with momentum to solve Eq.~\ref{eq:opt}.  Here, the update equation is given by 
\begin{equation}
\mathbf{x}^{j+1} = \mathbf{x}^{j} +\mathbf{v}^j,
\end{equation}
where the velocity term is
\begin{equation}
\mathbf{v}^{j+1}=\mu \mathbf{v}^j - \eta \nabla_{\mathbf{x}} J(\mathbf{x}^j).
\label{eq:momentum}
\end{equation}
The parameters $\mu$ and $\eta$ denote the momentum and learning rates, respectively.  We fixed these parameters as $\mu=0.9$ and $\eta=0.004$. The instantaneous gradient term in Eq.~\ref{eq:momentum}, which guides the solution towards better minima, is
\begin{align}
\begin{split}
  \nabla_{\mathbf{x}} & J(\mathbf{x}) = \sum_i \omega_i \nabla_{\mathbf{x}} J_i(\mathbf{x}) \\
  = & \left\{\begin{matrix}
  \sum_i \omega_i \nabla_{\mathbf{x}} \mathcal{L}(\tilde{\mathbf{x}}_i) &,\; \tilde{\mathbf{x}}_i = \{ x_{u,v}  : (u,v) \in \beta_i \}\\ 
 0 &,\; otherwise
\end{matrix}\right. .\\
\end{split}
\label{eq:gradient}
\end{align}
From Eq.~\ref{eq:gradient}, we see that the overall gradient is given by a linear combination of gradients of the regional objective functions: $\nabla_{\mathbf{x}} \mathcal{L}(\tilde{\mathbf{x}}_i)=\nabla_{\mathbf{x}} \ell(g(\tilde{\mathbf{x}}_i), h_i \circ g(\tilde{\mathbf{t}}_i)) + \lambda \nabla_{\mathbf{x}} R(\tilde{\mathbf{x}}_i)$. 

\begin{figure}[tbp]
\centering
\includegraphics[width=0.25\textwidth]{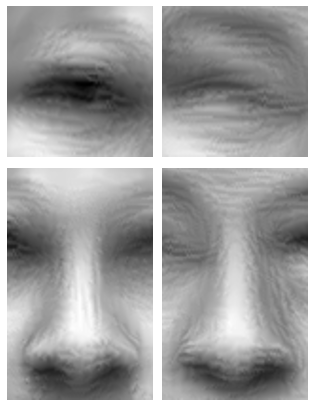}
\caption{Example synthesized imagery using a smaller, local regions (left column) versus using a larger, global region (right column).}
\label{fig:regions}
\end{figure}

Fundamentally, there are two ways the different regions may affect the synthesis results.  The first is through the distinctive thermal-to-visible mapping $h_i$, which is trained using only features from the corresponding region.  In other words, the mapping for the left eye is trained using only exemplars from the left eye region; the mapping for the entire face uses exemplars from across the face; and so on.  The second way is through the regularization penalties, especially the total variation component.  Given that the total variation defined as the sum of gradient magnitudes over a given set of pixels, there is a direct dependence on the size of the image (or region).  Thus, a larger region (e.g., the entire face region) is penalized more than a smaller region, yielding an image that blurs local details.  This concept is illustrated in Fig.~\ref{fig:regions}.  

\subsection{Implementation Details}
Until now, we have ignored the specific architectures used for the mappings $g(\cdot)$ and $h_i(\cdot)$ for $i=1\dots K$.  

The input images used in our implementation are the $S_0$ images for the conventional thermal case and the $S_0$ and DoLP images for the polarimetric case.  Moreover, when training, we augment the input images by applying a Non-Local Means (NLM) photometric normalization filter \cite{Struc2010}.  The output targets are the extracted feature from $g(\cdot)$ using corresponding visible images.  Note that no NLM normalization is applied to the output.  We augment the training data in order to alleviate potential issues with over-fitting.

The generic mapping, $g(\cdot)$, is a CNN that produces a set of feature maps.  Here, we assume that the types of features (e.g., HOG, SIFT, or some arbitrary deep feature representation) being extracted are identical between the visible and thermal images.  In \cite{MahendranVedaldi2015}, separate CNN architectures are defined to produce HOG and DSIFT feature maps for purposes of image understanding.  Here, we repurpose the DSIFT architecture for cross-spectrum synthesis. 

Next, the cross-spectrum mappings, $h_i(\cdot)$ for $i=1\dots K$ are all implemented as a $1\times1$ convolutional network and are trained to map features extracted from thermal images to the features from corresponding visible images.  Each of the these networks has two hidden layers with 200 hidden units and use the hyperbolic tangent activation function: $\sigma(u)=(1-\exp(-2u))/ (1+\exp(-2u))$.  
Thus, given corresponding pairs of visible and thermal features, $(y^{(k)}, z^{(k)})$ for $k=1\dots N$, which are computed by mapping corresponding images with $g(\cdot)$, we train the cross-spectrum mapping to minimize    
%While the synthesized imagery is impacted by the regularization penalties from various regions, the cross-spectrum component is primarily guided by the regional mappings $g_i(\mathbf{t})$.  

%Each mapping $g_i(\mathbf{t})$ is define as a CNN that is decomposed into two components: {\it feature extraction} and {\it feature regression}.   Here, we assume that the type of features being extracted between the visible and thermal imagery are the same (e.g., HOG, SIFT, or some other deep feature representation).  Therefore, we have 
%\begin{equation}
%g_i(\mathbf{t})=h_i \circ f(\mathbf{t})
%\end{equation}
%where $h_i(\mathbf{u})$ denotes the feature regression component, which is defined using $1\times1$ convolutions.  The mapping $g_i(\mathbf{t})$ is trained to minimize the sum square error, 
\begin{equation}
\sum_k \left \| y^{(k)} - h_i(z^{(k)})\right \|^2 
\end{equation}
for each region.

%between the visible feature maps and the transformed thermal feature maps.  In Eq. 7, $y^{(k)}$ and $g_i^{(k)}$ represent the $k^th$ feature map response vectors for the visible and mapped thermal inputs, respectively.  Note that this optimization is performed for each of the K local regions.  Thus, in addition to the multiple regularizations, our model also provides multiple estimates from which to synthesize a corresponding visible image.

After training $h_i(\cdot)$ for $i=1\dots K$, the multi-region optimization in Eq.~\ref{eq:opt} can be performed.  We adapted the source code from \cite{MahendranVedaldi2015} to incorporate multiple predefined regions for synthesizing a visible image from a given thermal image.  This is done by tracking the gradient updates for each region and weighting  them accordingly.  Specifically, within the left and right eyes regions, the locally computed  gradients are weighted by a factor of 0.95 and the globally computed gradients are weighted by 0.05.  Within the region covering the nose and mouth, the locally computed gradients are weighted by a factor of 0.75 and the globally computed gradients are weighted by 0.25.  The remaining parts of the image only consider the globally computed gradients.

The biggest difference between this work and \cite{RigganShort2016b} is the generalization to multi-region cross-spectrum synthesis, which is shown to improve the quality of the synthesized imagery for matching against visible face imagery using existing face recognition algorithms (Section~\ref{sec:exp}).  

\subsection{Relevance to GANs}
Unlike the original GAN framework introduced by Goodfellow et al.~\cite{GoodfellowPouget2014}, which   fundamentally learns how to produce a random, but photo-realistic image from some underlying distribution, our multi-region cross-spectrum synthesis method aims to produce a visible image given a thermal image that is sufficiently discriminative---meaning that the synthesized image can be matched against gallery images with a high degree of accuracy.  Conceptually, this is not all that different from conditional GAN-based methods, like \cite{IsolaZhu2017}, where an image in one domain is generated based on an input from another domain.  However, the key objective for GAN-based methods is to produce imagery with similar photometric properties (e.g., dynamic range) as the underlying distribution (or conditional distribution) provided by the training set.  Thus, it is possible that GANs may not emphasis discriminability as much as  photo-realism.  Zhang et al.~\cite{ZhangPatel2017} alleviate the lack of discriminability, to some degree, by incorporating an identity loss as part of their GAN-based approach for cross-spectrum synthesis.  Whereas, our approach, which  is based on synthesizing images from edge-based features (both locally and globally), preserves much of the discriminative facial structure that is useful for matching faces.  In the following section, we demonstrate that our multi-region cross-spectrum synthesis method produces more discriminative imagery than current state-of-the-art approaches.  

\section{Experiments and Results}
\label{sec:exp}
We evaluate our multi-region cross-spectrum synthesis model (Section 3) using the database\footnote{The dataset is available to the research community upon request with a signed database release agreement.  Requests for the database can be made by contacting Shuowen (Sean) Hu at {\it shuowen.hu.civ@mail.mil}.} from \cite{HuShort2016}, which contains corresponding polarimetric thermal and visible faces from 60 distinct individuals. The imagery contains distance and expression variations. For complete information with regards to the data collection please refer to \cite{HuShort2016}.

For the experiments described in this paper, we use 30 subjects (baseline and expression) for training, and the remaining 30 subjects for evaluation.  The imagery is aligned using the eye coordinates, as in \cite{RigganShort2016b}.

For this dataset, we define the bounding boxes for the local regions to be: $[30\; 89\; 64\; 34]$ for the right eye region, $[106\; 89\; 64\; 34]$ for the left eye region, and $[70\; 125\; 65\; 85]$ for the region covering the nose and mouth.  Note the bounding box format is {\it upper left x coordinate}, {\it upper left y coordinate}, {\it width},  and {\it height}.  These defined regions are used with registered images from \cite{HuShort2016}, which are $200 \times 250$ pixels. 

We consider two use cases for our synthesis method: (1) adjudication of matches by an analyst and (2) utilization of existing face recognition components or systems (e.g., landmark detection and matching).  Therefore, we compare the resulting synthesized imagery from our approach with the state-of-the-art cross-spectrum synthesized imagery using: 

\begin{enumerate}
\item qualitative analysis,
\item verification performance,
\item landmark detection accuracy.
\end{enumerate}
This analysis highlights the key differences in visual appearance, discriminability, and interoperability of synthesize imagery.

\begin{figure*}[htbp]
  \centering
   \begin{subfigure}[b]{0.02\linewidth}
   \rotatebox{90}{\hspace{0.08cm} Polar-to-Vis \hspace{0.78cm} $S_0$-to-Vis} \\ \\
  \end{subfigure}
  ~
  \begin{subfigure}[b]{0.14\linewidth}
    \includegraphics[width=\linewidth]{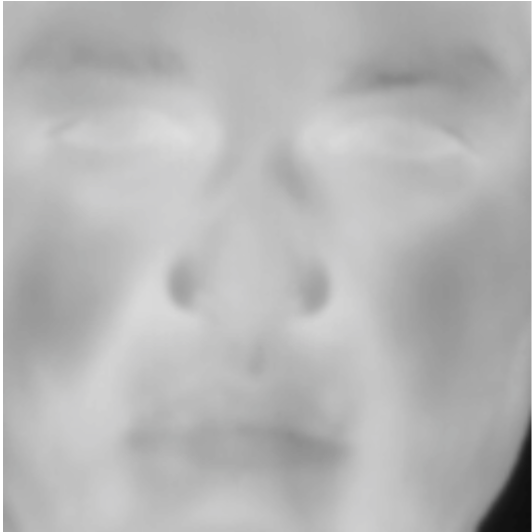}\\
    \includegraphics[width=\linewidth]{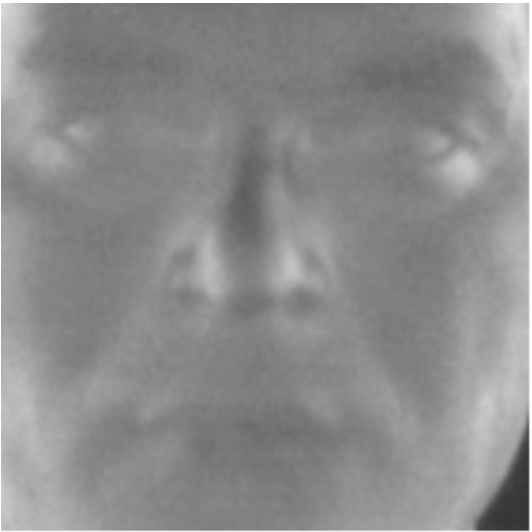}
    \caption{Raw}	
  \end{subfigure}
  ~
  \begin{subfigure}[b]{0.14\linewidth}
    \includegraphics[width=\linewidth]{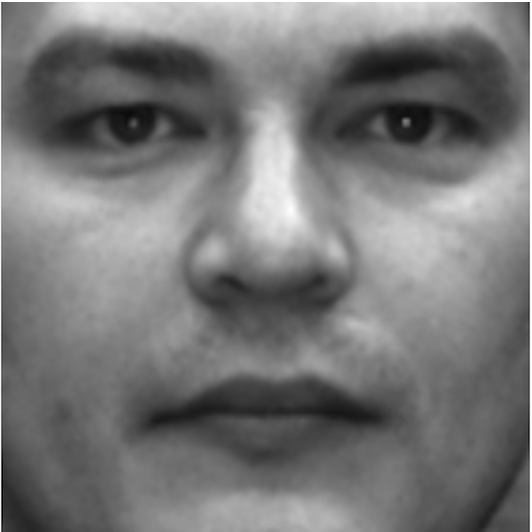}\\
    \includegraphics[width=\linewidth]{visible_gt.png}
    \caption{Ground Truth}	
  \end{subfigure} 
  ~
 \begin{subfigure}[b]{0.14\linewidth}
    \includegraphics[width=\linewidth]{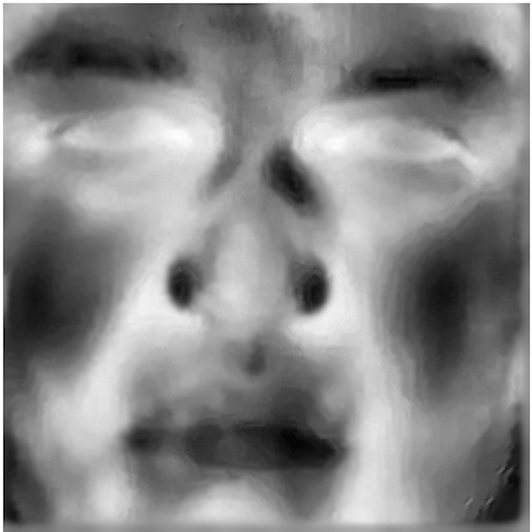}\\
    \includegraphics[width=\linewidth]{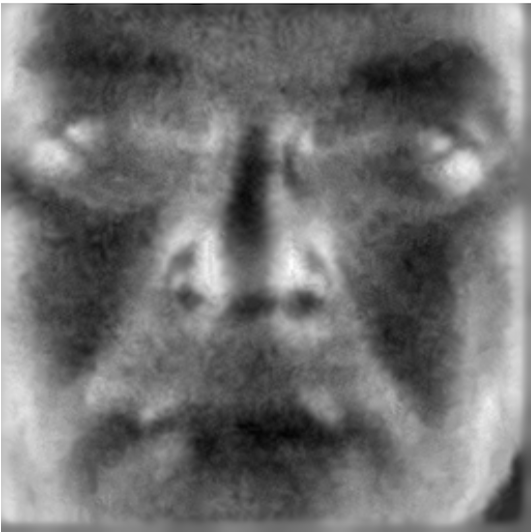}
    \caption{\cite{MahendranVedaldi2015}}
  \end{subfigure}
  ~
  \begin{subfigure}[b]{0.14\linewidth}
    \includegraphics[width=\linewidth]{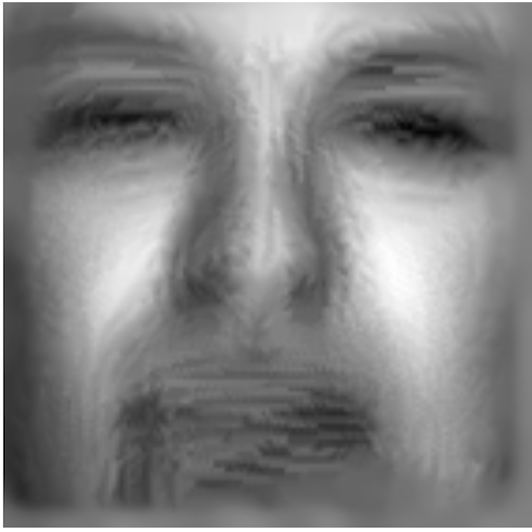}\\
    \includegraphics[width=\linewidth]{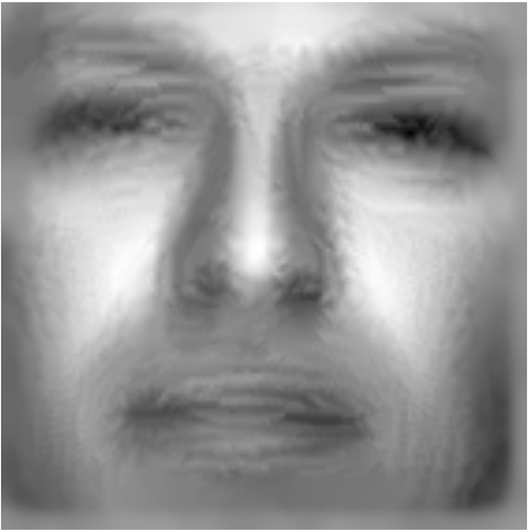}
    \caption{\cite{RigganShort2016b}}
  \end{subfigure}
  ~
   \begin{subfigure}[b]{0.14\linewidth}
    \includegraphics[width=\linewidth]{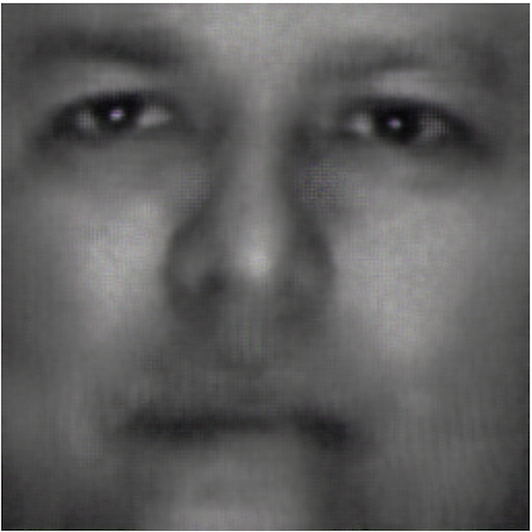}\\
    \includegraphics[width=\linewidth]{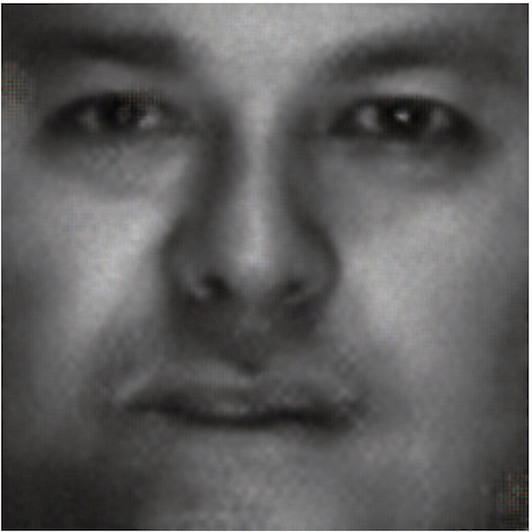}
    \caption{\cite{ZhangPatel2017}}
  \end{subfigure}
  ~
   \begin{subfigure}[b]{0.14\linewidth}
    \includegraphics[width=\linewidth]{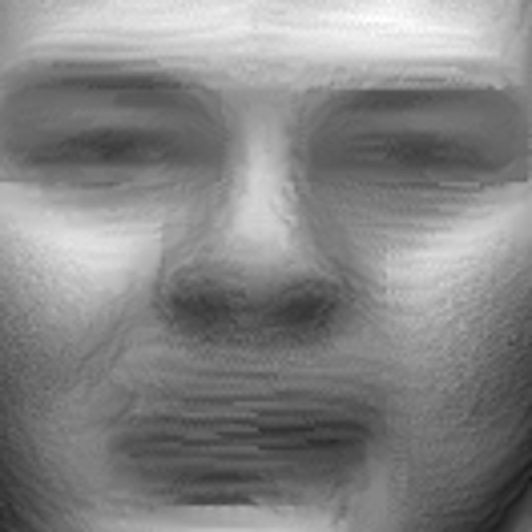} \\
    \includegraphics[width=\linewidth]{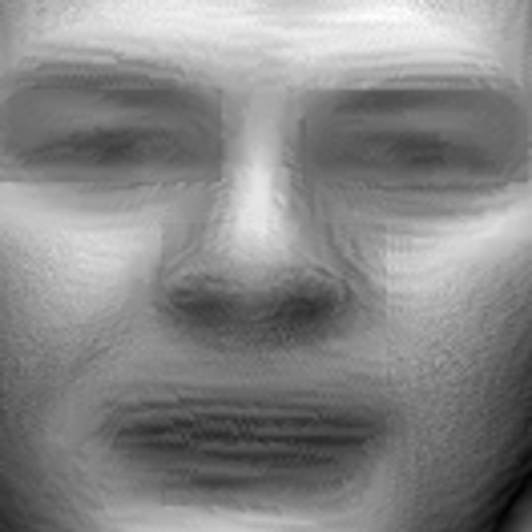}
    \caption{{\bf ours}}
  \end{subfigure}
%  \begin{subfigure}[b]{0.4\linewidth}
%    \includegraphics[width=\linewidth]{coffee.jpg}
%    \caption{More coffee.}
%  \end{subfigure}
  \caption{Comparison of synthesis methods.}
  \label{fig:results1}
\end{figure*}

\subsection{Qualitative Analysis}
In Fig.~\ref{fig:results1}, we compare multiple thermal-to-visible synthesis based approaches.  The baseline approach \cite{MahendranVedaldi2015}, which does not explicitly perform cross-spectrum synthesis, produces imagery akin to thermal imagery.  By comparing the results \cite{RigganShort2016b} with our results using multiple local regions, the benefits of enhanced geometric and fiducial details are observed.  Also, when comparing our results with the photo-realistic results from \cite{ZhangPatel2017}, it is not clear whether the photo-realistic texture from synthesis is a critical requirement for making visual comparisons between a visible image and a synthesized.  Although the texture produced by our approach is not photo-realistic, it does produce images that are more structurally similar to the ground truth visible images, which is important for discriminability (both for humans and for recognition algorithms).

\begin{figure*}[htb]
\centering
\includegraphics[width=0.60\linewidth]{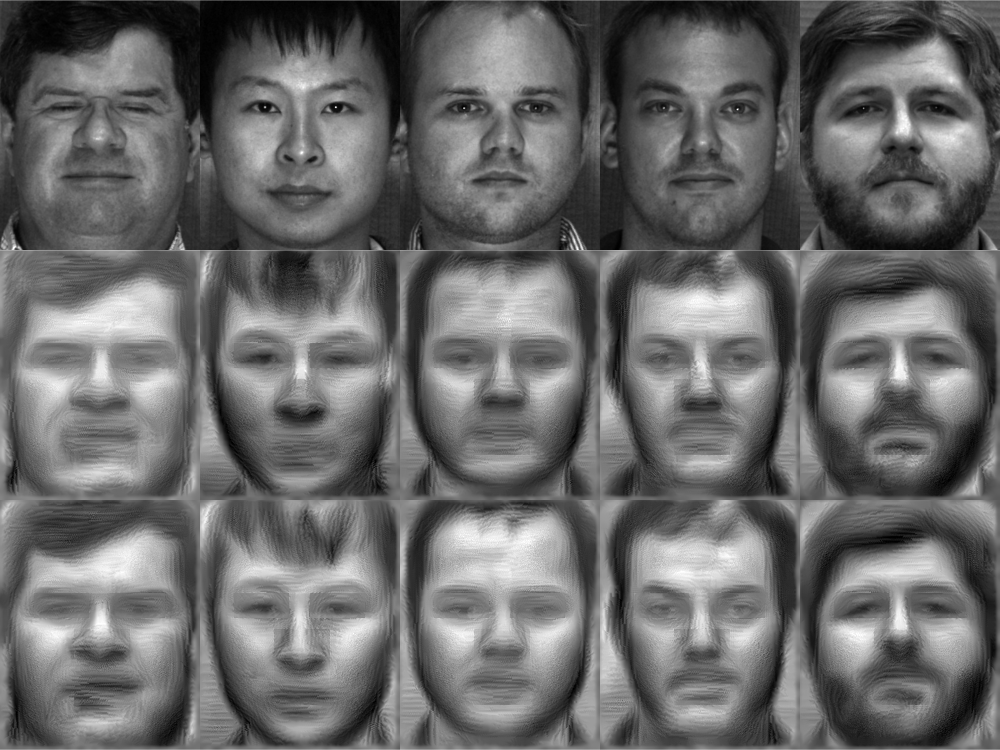}
\caption{Discriminability of ground truth images (top row), $S_0$-to-Vis synthesize imagery (middle row), and Polar-to-Vis synthesized imagery (bottom row)}
\label{fig:results2}
\end{figure*}

Additionally, we compare the synthesized imagery of multiple subjects side-by-side (Fig.~\ref{fig:results2}) in an attempt to qualitatively assess the discriminative nature of synthesis methods .  Although the former approach does a better job at synthesizing the appearance of the visible spectrum imagery, our approach seems to capture more of the structural information, which is often useful for face recognition.

The apparent synthesis artifacts in our approach (as well as in \cite{RigganShort2016b}) are partially due to known issues related to using a total variation based reconstruction error term, as pointed out in \cite{MahendranVedaldi2015}. These artifacts may be enhanced even further by errors introduced from cross-spectrum mapping.  However, despite such artifacts and lack of photo-realism, we demonstrate in section~\ref{sec:verfication} that our multi-region synthesis achieves better verification performance than state-of-the-art methods.  This may not be so surprising since deep neural networks are often trained to tolerate specific natural variations in facial appearance, such as changes in pose, illumination, and expression.

\subsection{Verification Performance}
\label{sec:verfication}
Next, we compare the Area Under the Curve (AUC) and and Equal Error Rate (EER) measures from the Receiver Operating Characteristic (ROC) curves when matching various types of synthesized faces with visible gallery images.  For fair comparison, we perform matching using the same tightly cropped regions as in \cite{RigganShort2016b,ZhangPatel2017}. Table~\ref{tab:verification} reports the AUC and EER for baseline methods, state-of-the-art approaches, and our new multi-region thermal-to-visible synthesis method.  Note that all methods use the vgg-face architecture \cite{ParkhiVedaldi2015}, where the last fully connected layer, prior to the softmax classifier, is used to compute a latent representation for gallery and probe images.  Then, we use the cosine similarity measure to produce similarity scores between latent representations from gallery and probe images.

In most cases, we use the pre-trained vgg-face model, which is trained using only visible faces, to demonstrated the discriminability of our synthesized imagery.  The only exception is that the ``Raw'' method in Table~\ref{tab:verification} is fine-tuned with thermal facial imagery, which demonstrates that transfer learning may not be able to overcome such a large modality gap for thermal-to-visible face recognition.

The baseline methods: ``Raw''  and Manhendran et al.~\cite{MahendranVedaldi2015} serve to demonstrate
%The two baseline methods are (1) the vgg-face model, fine-tuned with thermal facial imagery, and (2) the synthesis approach from \cite{MahendranVedaldi2015} which uses the pretrained vgg-face model \cite{ParkhiVedaldi2015}.  
%The former baseline demonstrates 
the need for a synthesis-based approach 
and
%while the latter baseline method highlights 
a cross-spectrum component, respectively.  
We also compare our multi-region synthesis approach with \cite{RigganShort2016b}, which is a special case of this work (i.e., a single region), to demonstrate how using multiple local regions to synthesize a more visible-like image impacts verification performance. Lastly, we compare our multi-region synthesis method with the state-of-the-art performance in Zhang et al.~\cite{ZhangPatel2017}. 

From the table, we see that our multi-region synthesis method achieves about a 5\% improvement in AUC over the state-of-the-art when using polarization state information, and about a 3\% improvement in AUC over the state-of-the-art for conventional thermal-to-visible synthesis.  Also, an improvement of about 4\% in EER over the state-of-the-art is observed, and about 1\% improvement in EER over the state-of-the-art for the conventional thermal case.

%Additionally, in Figure~\ref{}, we compare ROC curves between using a single region and multiple regions for both polarimetric and conventional thermal cases.  Here, we observe a notable improvement when combining both local and global region using our synthesis approach.

\begin{table*}[htpb]
\caption{Verification performance comparison between baseline methods, state-of-the-art methods, and our multi-region synthesis method for both polarimetric thermal (polar) and conventional thermal ($S_0$) cases.}
\begin{center}
\begin{tabular}{|c|c|c|c|c|}
 \hline 
 Method & AUC (polar) & AUC ($S_0$) & EER (polar) & EER ($S_0$) \\ \hline
 Raw & 50.35\% & 58.64\% & 48.96\% & 43.96\% \\
Mahendran et al.~\cite{MahendranVedaldi2015} & 58.38\% & 59.25\% & 44.56\% & 43.56\% \\ \hline
Riggan et al.~ \cite{RigganShort2016b} & 75.83\% & 68.52\% & 33.20\% & 34.36\% \\
Zhang et al.~\cite{ZhangPatel2017} & 79.90\% & 79.30\% & 25.17\% & 27.34\% \\ \hline
 Multi-Region Synthesis ({\bf ours}) & {\bf 85.43\%} & {\bf 82.49\%} & {\bf 21.46\%} & {\bf 26.25\%} \\ \hline
 
\end{tabular}
\end{center}
\label{tab:verification}
\end{table*}%

\subsection{Landmark Detection}
Lastly, we consider evaluating the accuracy of using our synthesize imagery for landmark detection of thermal imagery.  Often, face recognition performance depends upon the quality of landmark detection.  For thermal-to-visible face recognition, this is a challenging task since few (e.g., \cite{WangLiu2013}) have focused on landmark detection for thermal images.  Therefore, we address this problem by evaluating the accuracy of the detected landmarks in our synthesize imagery.  We perform this experiment by using pre-aligned visible and thermal imagery.  68 landmarks from the visible imagery are used as ``ground truth,'' which are obtained using DLIB's \cite{dlib09} landmark detector.  Then, we extract the same 68 landmarks from the synthesize imagery and report the average Euclidean distance between the two sets of landmarks. We found the landmark errors to be {\bf 4.95} pixels and {\bf 4.83} pixels on average for synthesized images from given conventional and polarimetric thermal images, respectively.  Without synthesis, applying existing face and landmark detection software (designed for visible imagery) directly to thermal facial imagery is pointless, since few (if any) faces and landmarks are detected.  Therefore, cross-spectrum synthesis methods, such as ours or \cite{ZhangPatel2017}, can be used to enable cross-spectrum landmark detection without having to design custom detectors for thermal imagery.

\begin{figure}[htbp]
\centering
\includegraphics[width=0.22\linewidth]{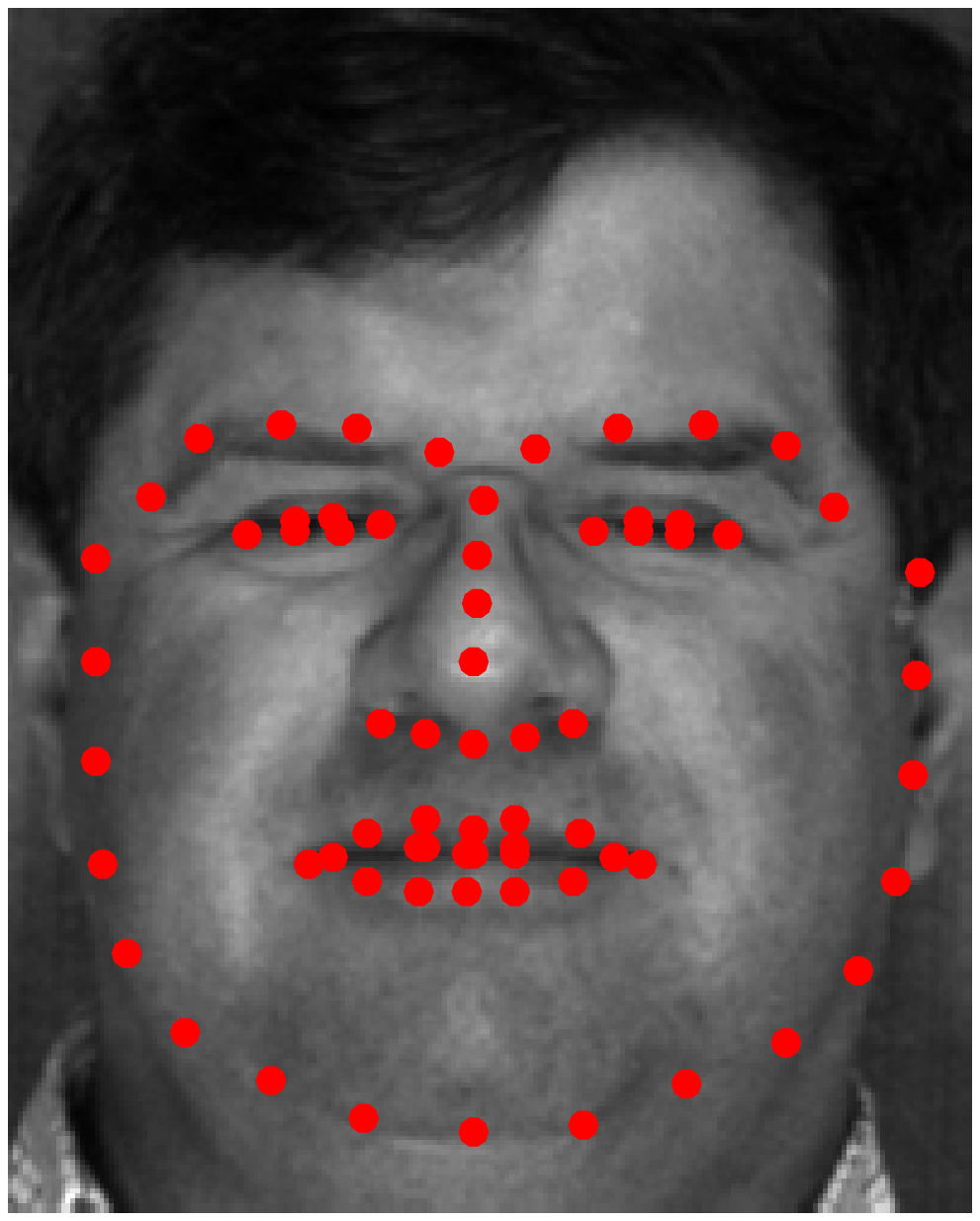}
\includegraphics[width=0.22\linewidth]{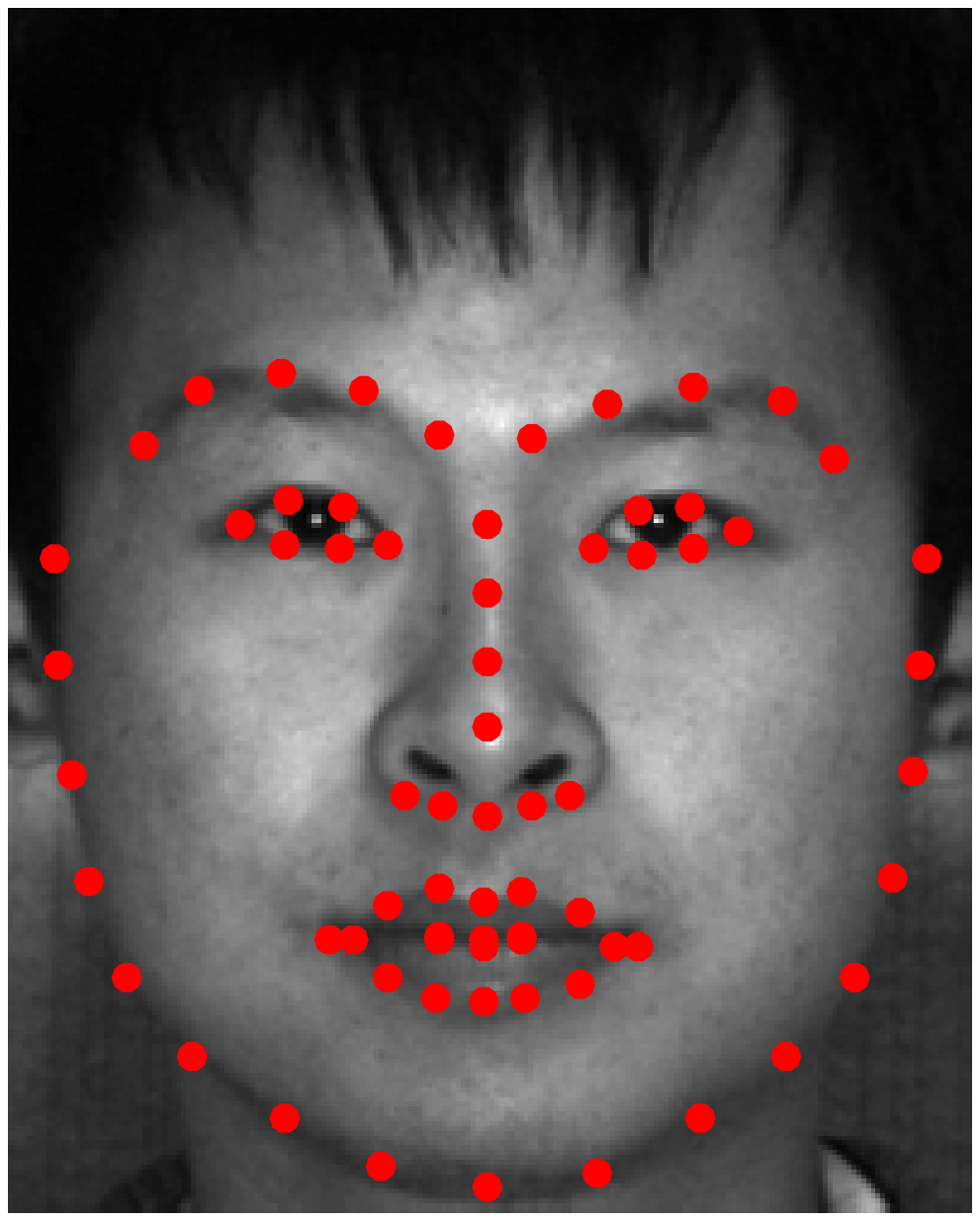}
\includegraphics[width=0.22\linewidth]{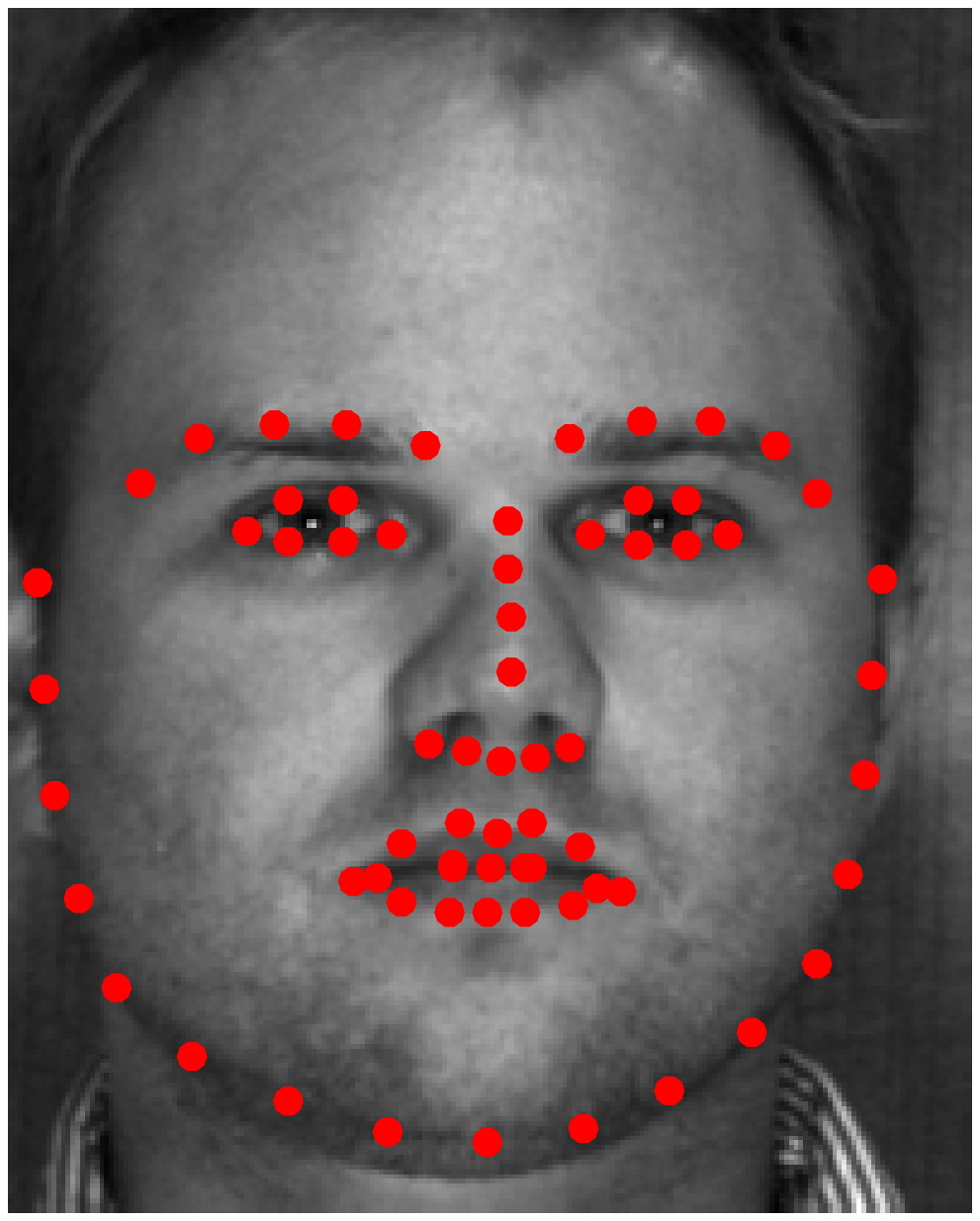}
\includegraphics[width=0.22\linewidth]{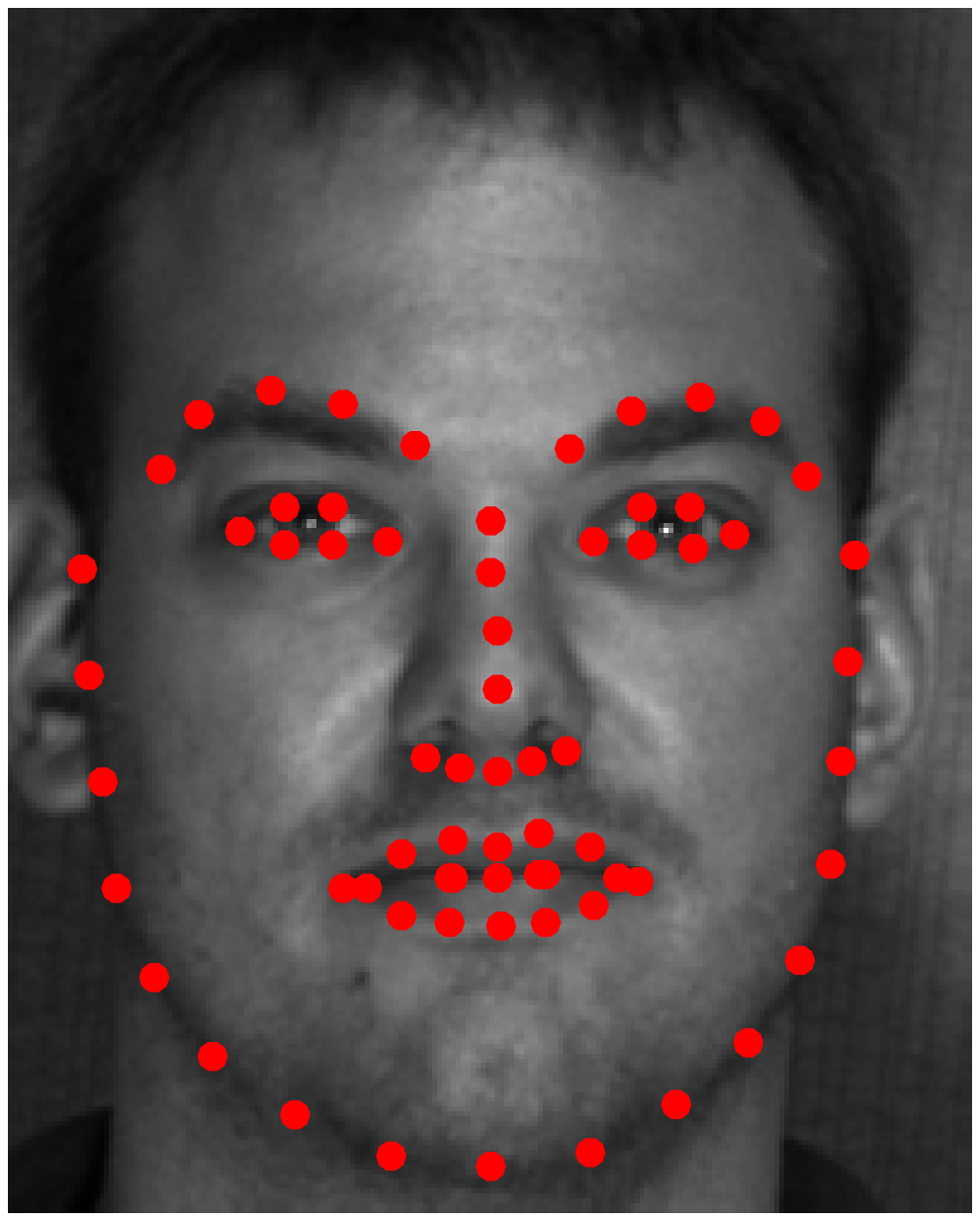}\\
\includegraphics[width=0.22\linewidth]{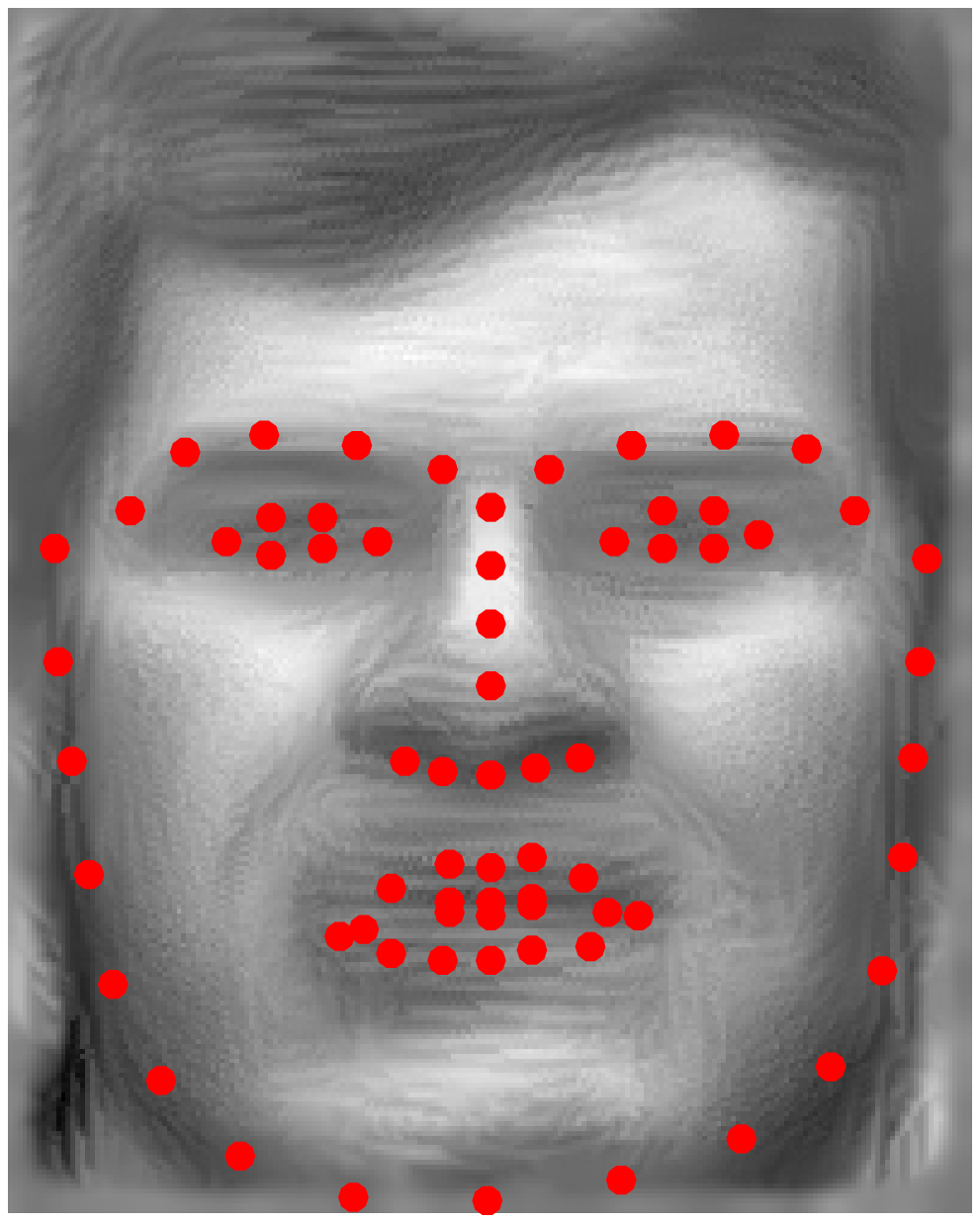}
\includegraphics[width=0.22\linewidth]{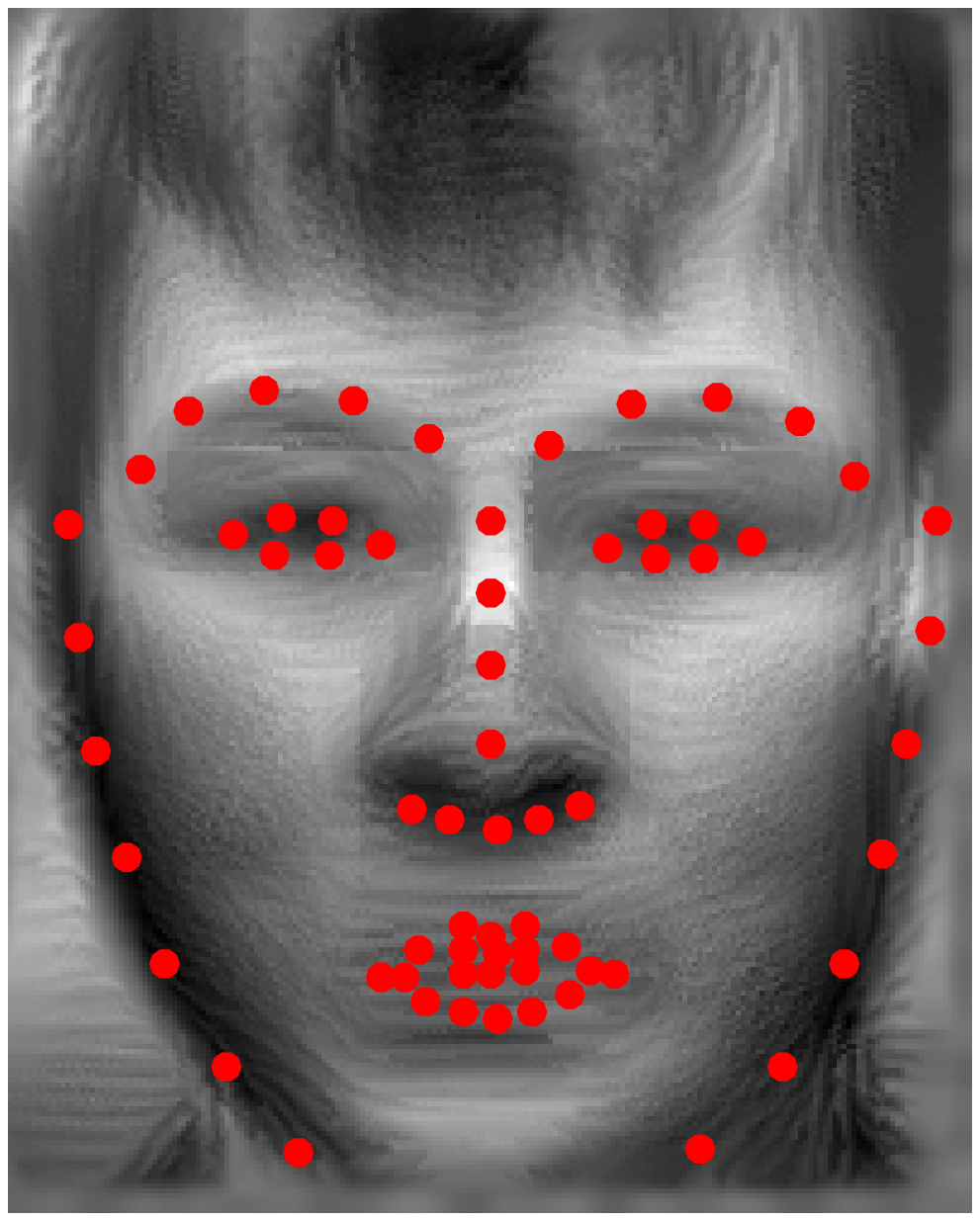}
\includegraphics[width=0.22\linewidth]{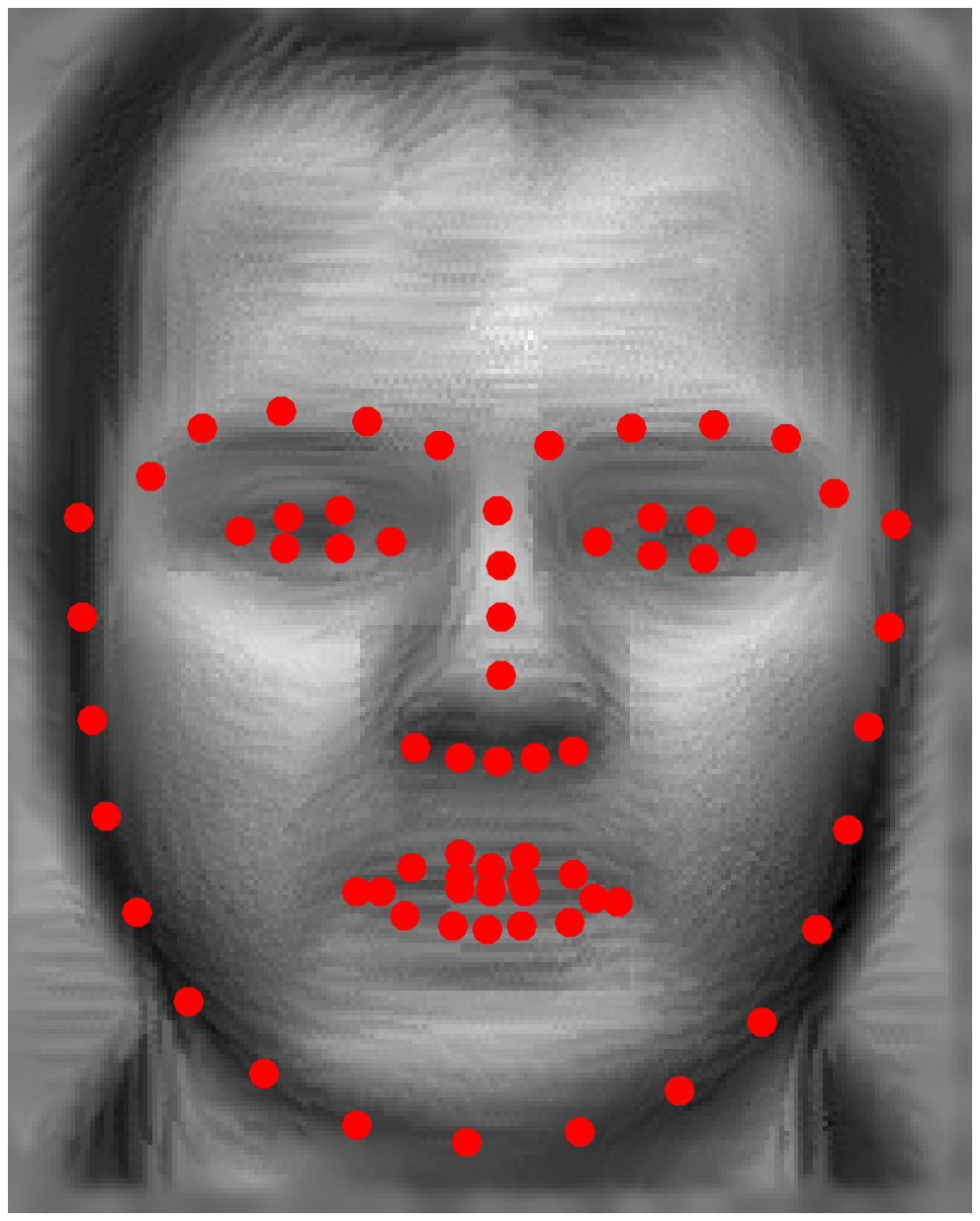}
\includegraphics[width=0.22\linewidth]{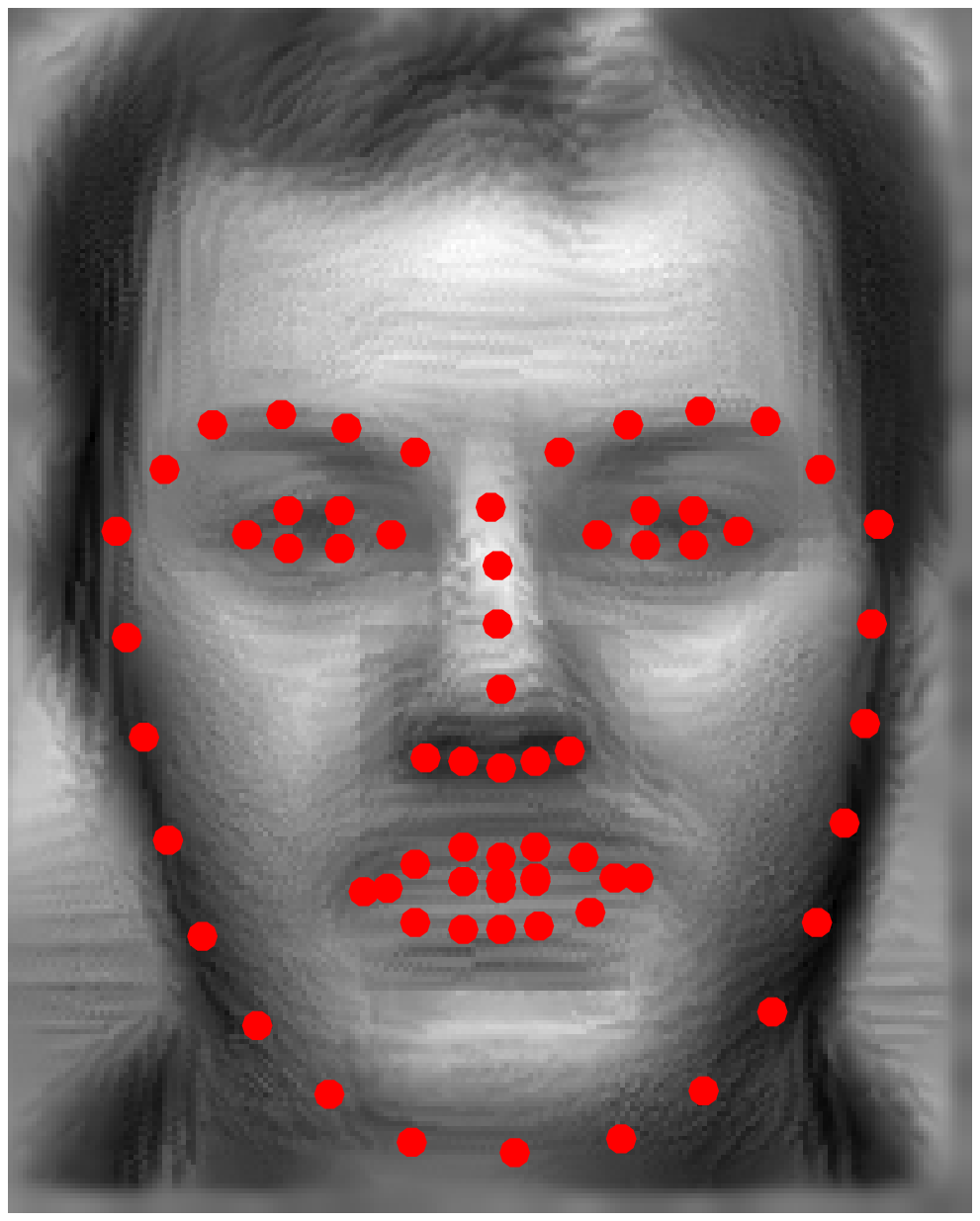}\\
\includegraphics[width=0.22\linewidth]{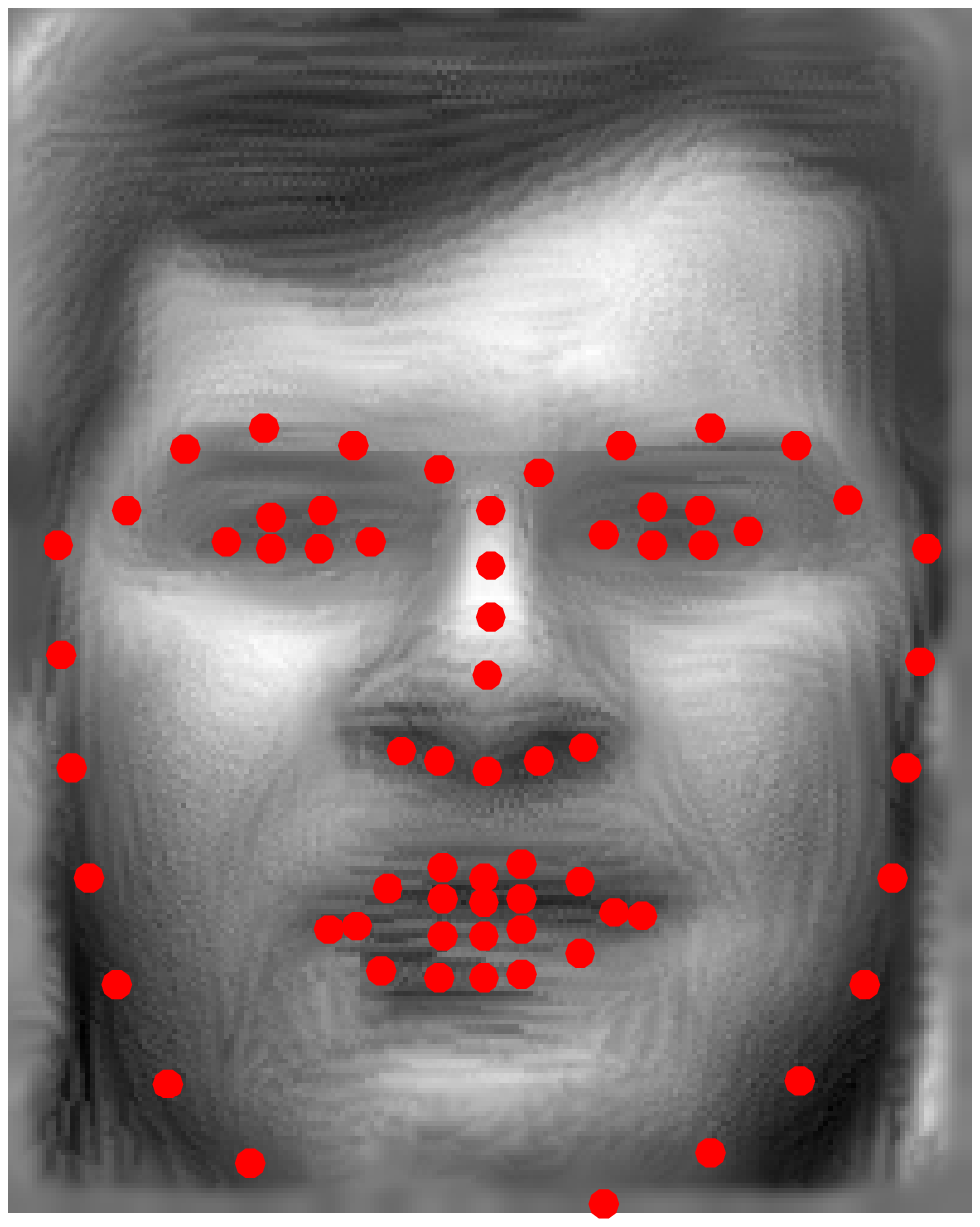}
\includegraphics[width=0.22\linewidth]{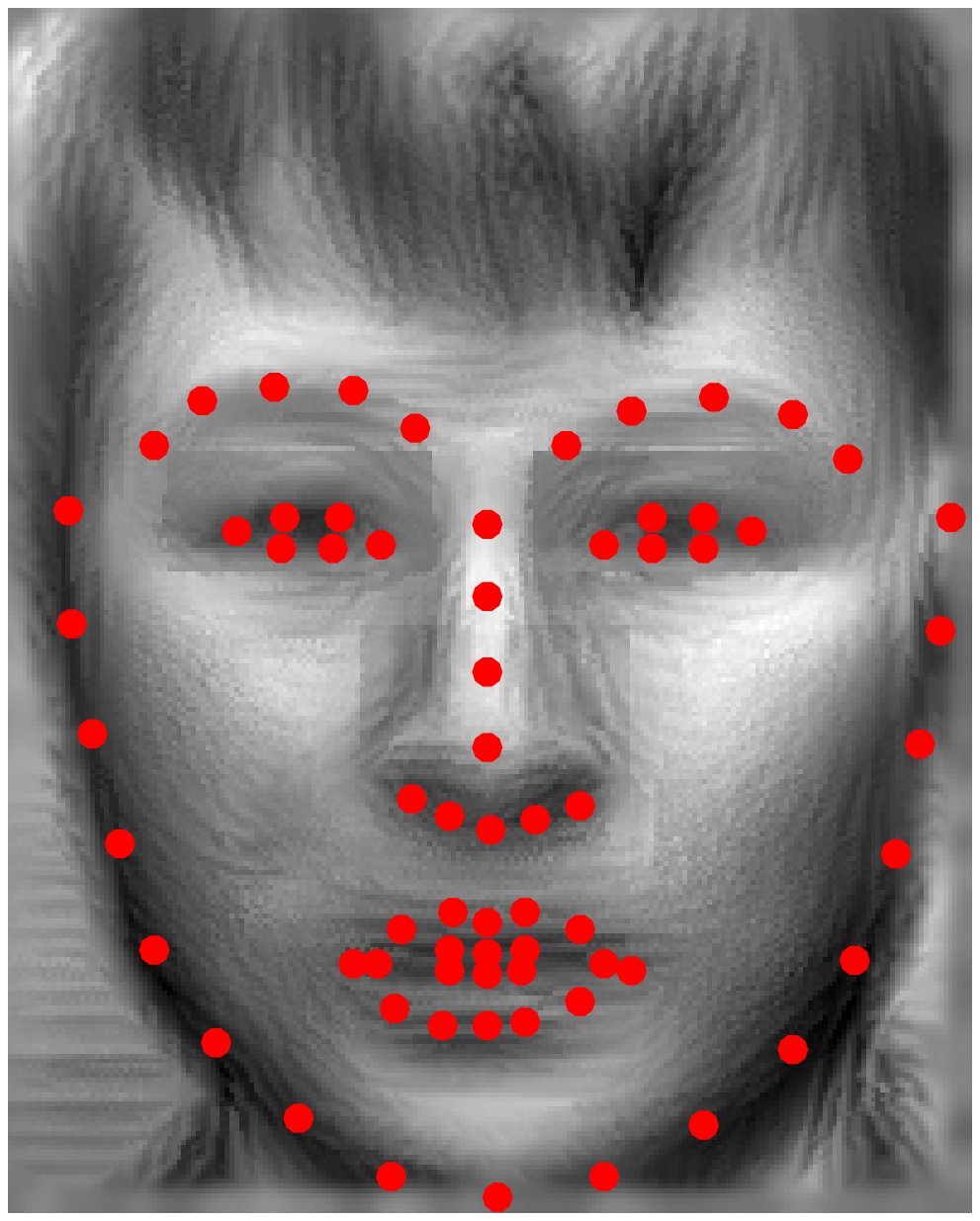}
\includegraphics[width=0.22\linewidth]{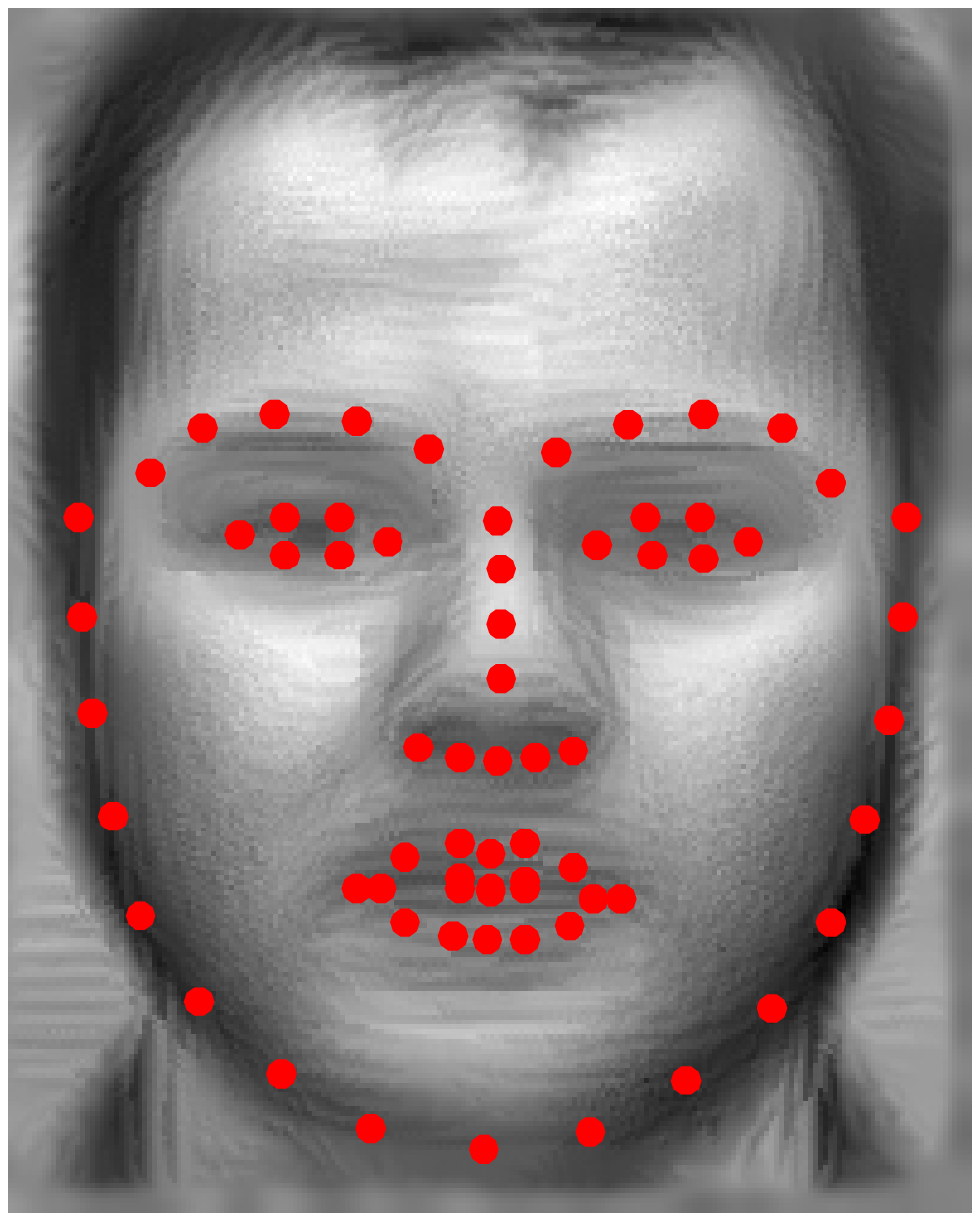}
\includegraphics[width=0.22\linewidth]{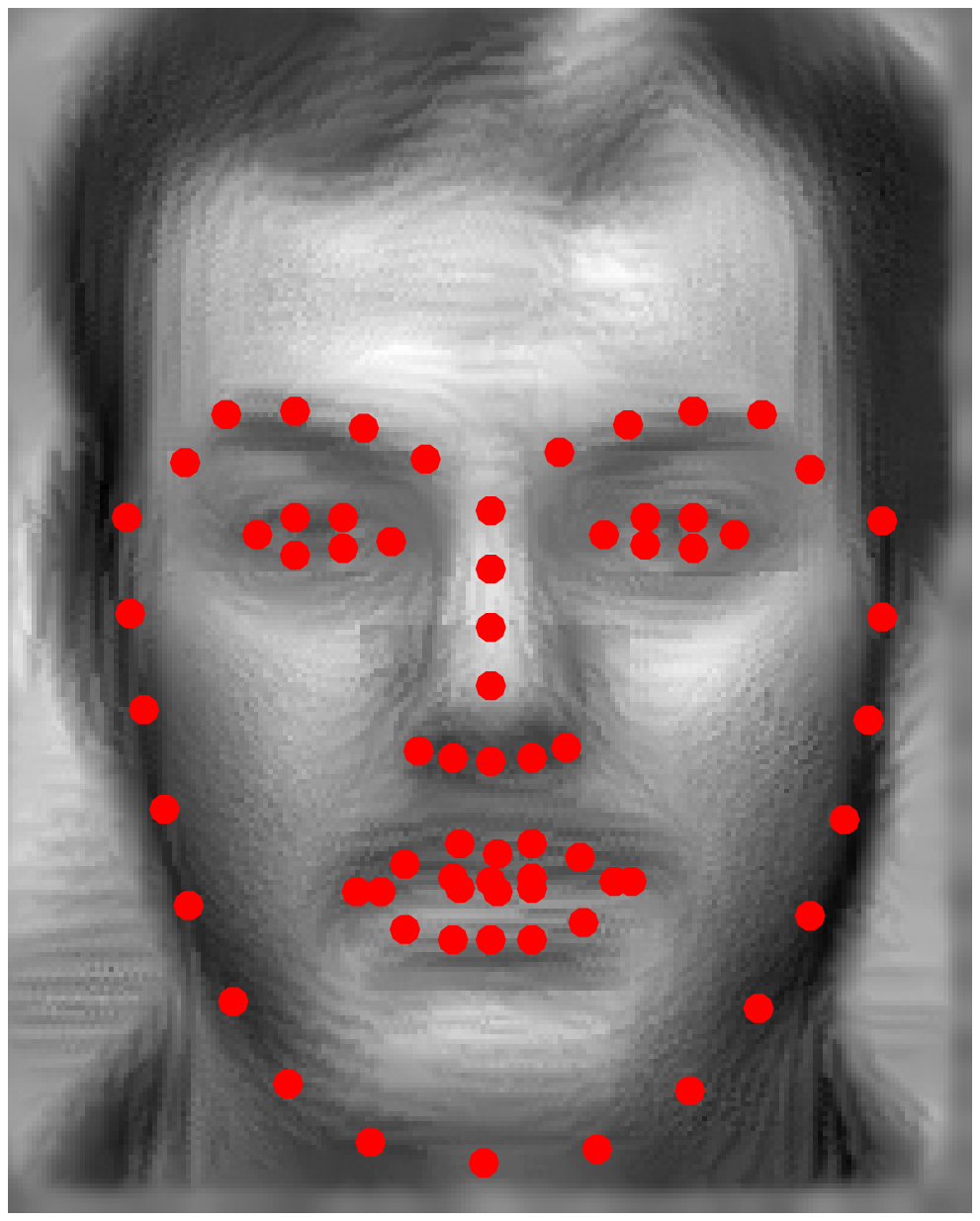}
\caption{Landmark detections for ground truth images (top row), $S_0$-to-Vis synthesize imagery (middle row), and Polar-to-Vis synthesized imagery (bottom row)}
\label{fig:landmarks}
\end{figure}

Fig.~\ref{fig:landmarks} shows several examples of aligned visible and thermal imagery with the detected landmarks.  This figure shows that we can detect the fiducial landmarks with a certain degree of accuracy (i.e., within a few pixels) when using the multi-region synthesize imagery, which can be helpful when matching thermal and visible faces.

%When evaluating face dete		ction using our synthesized imagery, we use the {\it failure-to-enroll} (FTE) as the measure for comparison.
%
%We compare the FTE rates between polarimetric multi- region synthesis (Polar-MR), conventional thermal multi- region synthesis, polarimetric baseline, and S0 baseline methods for vgg-face, PittPatt, and Neurotech face detectors. The results clearly show that the proposed multi-region synthesis method significantly improves performance of the various face detectors compare to using the original thermal or polarimetric thermal imagery.

%------------------------------------------------------------------------
\section{Conclusions}
In this paper, we presented a new multi-region synthesis approach that produces discriminative visible-like faces from either conventional thermal or polarimetric thermal facial imagery.  We also demonstrated that our synthesized imagery provides new state-of-the-art verification performance, which better facilitates interoperability with existing visible face recognition frameworks, such as the open source vgg-face model.  Furthermore, our synthesized imagery was shown to be effective for accurately detecting landmarks in thermal imagery, which is a critical component of face recognition systems.

We showed that by jointly optimizing over both global and local facial regions, which provided complementary representations and diverse regularization penalties, we were able to produce highly discriminative representations.  Despite the fact that the synthesized imagery does not produce a photo-realistic texture, the verification performance achieved was better than both baseline and recent approaches when matching the synthesized faces with visible faces.

%We demonstrated that our multi-region synthesis approach improved COTS/GOTS and OS face recognition rank-1 ID rate over the single region synthesis approach (similar to \cite{RigganShort2016b}) by as much as ~6\% (Neurotech) for polarimetric thermal , and as much as ~11\% (vgg-face) for conventional thermal. Moreover, we demonstrated

Secondly, we demonstrated that the synthesized imagery provided by our multi-region optimization method can be used to enable landmark detection capabilities for conventional thermal or polarimetric thermal imagery, where few landmark detection methods currently exist.  Since many facial recognition pipelines typically depend on accurate landmark detection, this proves to be beneficial for thermal-to-visible face recognition.  We showed that when using DLIB's \cite{dlib09} open-source landmark detection software on our synthesized imagery, the landmarks were detected accurately (less than 5 pixels in distance).

While recognition performance using custom cross-spectrum methodologies may be currently superior to those reported in this paper, the fundamental benefits of our framework include: the ability to adjudicate potential matches and the ability to leverage state-of-the-art visible spectrum face recognition technology for thermal-to-visible matching.

{\small
\bibliographystyle{ieee}
\bibliography{wacv_references}
}

\end{document}